\documentclass[10pt,a4paper]{article}
\usepackage[T1]{fontenc}
\usepackage[utf8]{inputenc}
\usepackage[a4paper,left=18.1mm,right=18.1mm,top=20mm,bottom=20mm,columnsep=18pt,footskip=14pt]{geometry}
\usepackage{newtxtext,newtxmath}
\usepackage{graphicx,amsmath}
\usepackage{booktabs,makecell,multirow,array,colortbl,dcolumn,threeparttable}
\usepackage{balance,stfloats}
\usepackage[numbers,sort&compress]{natbib}
\usepackage{enumitem}
\usepackage{titlesec}
\usepackage{xcolor}
\usepackage{caption}
\usepackage{hyperref}
\hypersetup{hidelinks,
  pdftitle={CompliantBot: Learning Whole-Body Compliance across Task Interfaces for Humanoid Control},
  pdfauthor={Haoyang Li, Keqi Zhu, Zhanxiang Cao, Yang Zhang, Yizhi Chen, Xuekang Yang, Yue Gao},
  pdfsubject={Author-prepared research preprint},
  pdfkeywords={Humanoids, Compliance, Reinforcement learning, Motion tracking, Physical interaction}}
\newcommand{\vect}[1]{\boldsymbol{#1}}
\newcommand{\CBcaptionfont}{\rmfamily\fontsize{8.5}{10}\selectfont}
\newcommand{\CBtablefont}{\rmfamily\fontsize{8.5}{10}\selectfont}

\DeclareCaptionFont{cbcaption}{\CBcaptionfont}
\titleformat{\section}{\normalfont\fontsize{12}{14}\selectfont\bfseries}{\thesection.}{0.5em}{}
\titleformat{\subsection}{\normalfont\fontsize{11}{13}\selectfont\bfseries}{\thesubsection.}{0.5em}{}
\titlespacing*{\section}{0pt}{10pt plus 2pt minus 2pt}{5pt}
\titlespacing*{\subsection}{0pt}{8pt plus 2pt minus 2pt}{4pt}
\setlist{leftmargin=1.5em,itemsep=2pt,parsep=0pt,topsep=4pt}
\graphicspath{{./}}
\begin{document}
\begin{center}
{\fontsize{18}{22}\selectfont\bfseries
CompliantBot: Learning Whole-Body Compliance\\
across Task Interfaces for Humanoid Control\par}
\vspace{9pt}
{\fontsize{11}{14}\selectfont
Haoyang Li\textsuperscript{1,4}, Keqi Zhu\textsuperscript{4}, Zhanxiang Cao\textsuperscript{2,4}, Yang Zhang\textsuperscript{1},\\
Yizhi Chen\textsuperscript{4}, Xuekang Yang\textsuperscript{2}, Yue Gao\textsuperscript{3,4,*}\par}
\vspace{6pt}
{\fontsize{8.5}{10.5}\selectfont
\textsuperscript{1}Department of Automation, Shanghai Jiao Tong University, Shanghai, China\\
\textsuperscript{2}Department of Computer Science, Shanghai Jiao Tong University, Shanghai, China\\
\textsuperscript{3}MoE Key Lab of Artificial Intelligence and AI Institute, Shanghai Jiao Tong University, Shanghai, China\\
\textsuperscript{4}Shanghai Innovation Institute, Shanghai, China\par}
\end{center}
\vspace{4pt}
\begin{center}\textbf{Abstract}\end{center}
{\small\noindent
Humanoid controllers need the ability to yield to external forces while retaining the capacity to resume their original tasks. However, tracking objectives penalize the deviations needed for yielding, while relaxing tracking goals persistently can make task recovery harder. Existing approaches often rely on task-specific interaction command interfaces or constructed compliant motion data and reference trajectories, limiting the direct reuse of their compliance-training mechanisms across different task interfaces. We introduce CompliantBot, a compliance-training framework for motion tracking and velocity-command locomotion. It preserves the original task inputs and requires no preconstructed compliant data or target trajectories. A state-dependent interaction force field couples applied loads to the robot's motion, enabling local articulation and whole-body movement to change the forces experienced. A complementary force-magnitude reward encourages lower interaction loads, while task rewards are downweighted during loading and restored after release to coordinate yielding and task resumption. The trained policies use task inputs and proprioceptive history without explicit external-force measurements. Across both task interfaces, CompliantBot yields farther under constant pulls and experiences lower spring interaction loads than matched baselines, while resuming task tracking after force removal. CompliantBot achieves the highest task completion among the compared methods under stronger pulls. Power evaluations further show that CompliantBot has the lowest mean joint mechanical power among the compared methods. Real-robot demonstrations further illustrate compliant responses through arm motion, stepping, and body reorientation.

\par}
\vspace{5pt}
{\small\noindent\textbf{Keywords:} Humanoids; Compliance; Reinforcement learning; Motion tracking; Physical interaction.\par}
\vspace{10pt}
\begingroup
\centering
\includegraphics[width=0.99\textwidth]{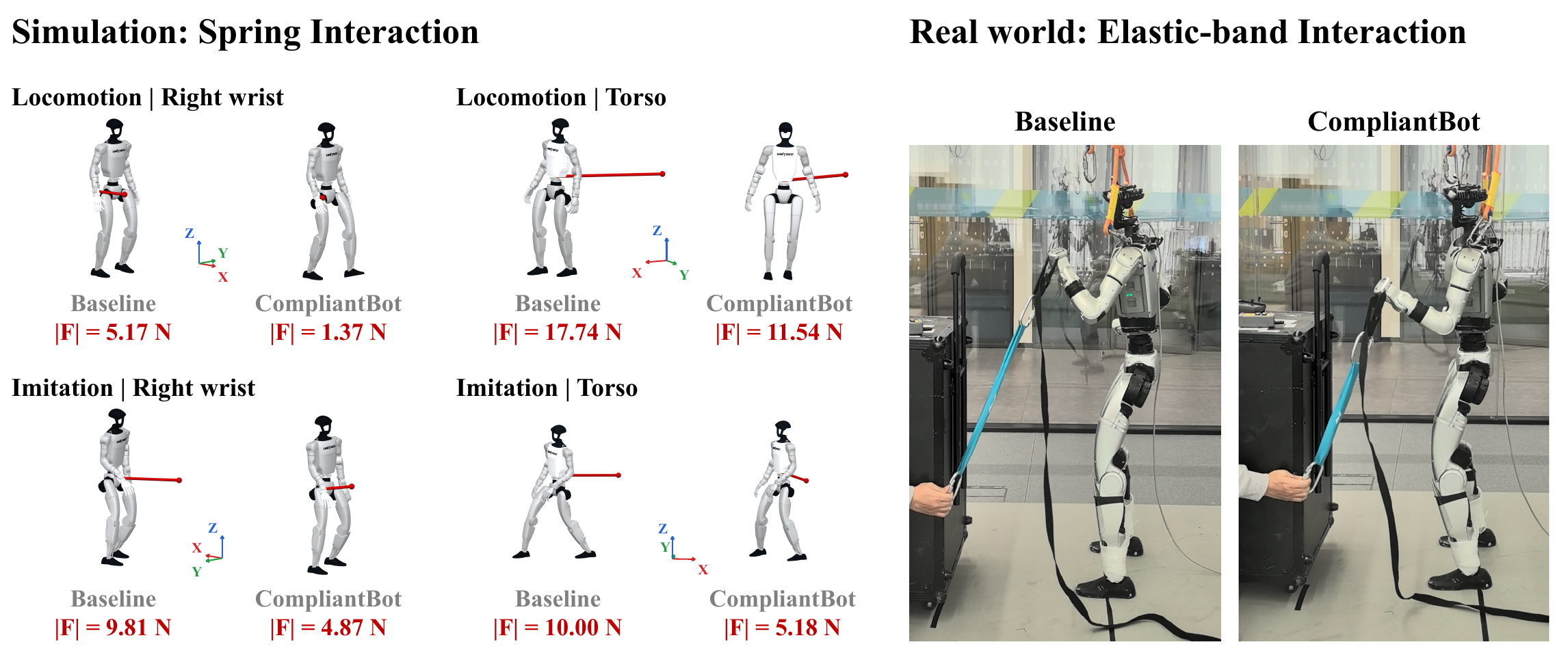}
\par\smallskip
\refstepcounter{figure}\label{fig:teaser}
\begin{minipage}{\textwidth}\CBcaptionfont
\textbf{Fig.~\thefigure. Compliant interaction in simulation and on hardware.} Left: Baseline and CompliantBot responses to fixed-anchor springs at the right wrist and torso during locomotion and motion imitation. Red lines indicate springs, and labels show instantaneous force magnitudes. Right: Elastic-band wrist interaction on hardware. These snapshots illustrate compliant responses to external forces through local and whole-body motion.
\end{minipage}

\par\endgroup
\begingroup
\renewcommand{\thefootnote}{\fnsymbol{footnote}}
\footnotetext[1]{Corresponding author: Yue Gao (\href{mailto:yuegao@sjtu.edu.cn}{yuegao@sjtu.edu.cn}).}
\endgroup
\setcounter{footnote}{0}
\clearpage
\twocolumn
\section{Introduction}
Learning-based humanoid whole-body control now spans locomotion, whole-body motion tracking, and loco-manipulation~\cite{beyondmimic,cassie-loco,gmt,twist,homie,hover}. Motion tracking lets a robot reproduce rich whole-body behaviors from reference motions~\cite{beyondmimic,gmt,twist}, while velocity-command controllers provide a complementary interface for robust walking and real-time response~\cite{cassie-loco}. Recent systems combine such modes into unified policies~\cite{homie,hover}. These advances largely prioritize accurate task execution and treat external effects as disturbances to reject. When these controllers move from the laboratory into spaces shared with people, however, they must also handle physical interaction: a person may guide the robot by its wrist, steady its torso, or redirect its motion through a gentle pull. Such interactions require the controller to balance task execution with responsiveness to external forces. Rejecting every displacement preserves task tracking but can sustain interaction forces; following a pull without restraint can compromise balance or lead the policy to abandon the task. A useful controller should therefore follow its motion reference or velocity command under nominal conditions, allow temporary deviations during interaction without losing balance, and resume the task after the external force is removed.

Prior work has shown that robots can learn compliant behavior from constructed compliant motion targets~\cite{softmimic,gentlehumanoid} or from changes in task priorities during interaction or balance recovery~\cite{deepcompliancego1}. These advances suggest that compliance depends not only on the forces encountered during training, but also on how motion under those forces is evaluated. During nominal execution, deviations from a motion reference or velocity command are errors to correct; during sustained physical interaction, temporary deviations can instead be necessary for yielding. Continually prioritizing task accuracy may discourage yielding to external forces, whereas persistently relaxing task tracking may compromise nominal performance and hinder task recovery after the external force is removed. For humanoids, this trade-off also involves coordination across the body: a pull at the wrist or torso may call for local joint motion, whole-body displacement, or both. Beyond this trade-off, different controllers expose different task interfaces, from whole-body motion references to velocity commands. A compliance-training approach applicable across different interfaces should therefore enable robots to yield to external forces and resume their tasks once the forces are removed. It should preserve the original motion reference or velocity command without requiring interface-specific compliant trajectories.

To this end, we introduce CompliantBot, a compliance-training framework for humanoid motion tracking and velocity-command locomotion that preserves the original task interface. Its central idea is to co-design a state-dependent interaction force field and learning objectives tailored to each interaction stage. During training, the field acts on sampled body links, allowing the robot to change the interaction forces through its own motion without a prescribed compliant trajectory. When external forces are applied, a force-magnitude objective encourages yielding, with the original task rewards downweighted to allow temporary task deviations. After the forces are removed, the original task weights are restored to encourage task recovery. The trained policy uses task inputs and proprioceptive history without requiring explicit external-force measurements.

Experiments show that CompliantBot achieves greater yielding to external forces than matched baselines while retaining the ability to resume the task after force removal. Results on both whole-body motion tracking and velocity-command locomotion demonstrate improved compliance, supporting the applicability of our approach across different task interfaces.

We make the following three contributions:
\begin{enumerate}
\item \textbf{Compliance training across task interfaces.}
We introduce CompliantBot, which enables compliant behavior in humanoid motion tracking and velocity-command locomotion while retaining the original task inputs, without requiring additional compliant target trajectories.
\item \textbf{A state-dependent interaction force field.}
We apply the interaction force field at sampled body sites, allowing the robot to regulate interaction forces through local articulation or whole-body motion.
\item \textbf{Reward allocation for yielding and task recovery.}
During training, a force-magnitude objective encourages yielding while the original task rewards are downweighted; their weights are restored after force removal to encourage task recovery. Regularization weights remain unchanged.
\end{enumerate}

\section{Related Work}
\subsection{Humanoid whole-body control}
Reinforcement learning (RL) supports multiple interfaces for humanoid control, including velocity-command locomotion, whole-body reference tracking, and loco-manipulation with separate upper- and lower-body commands. These interfaces specify complementary aspects of a task: desired movement through the environment, coordinated body motion, or manipulation combined with walking. Work on velocity-command locomotion addresses history-based adaptation~\cite{cassie-loco}, terrain-aware walking~\cite{h1-loco}, and smooth policy execution~\cite{lipschitz}. Li et al.~\cite{cassie-ijrr} further study dynamic bipedal skills using a dual-history policy architecture and task randomization, demonstrating adaptation to changes in dynamics and external disturbances. HuB~\cite{hub} addresses balance-intensive humanoid motions through reference refinement, balance-aware learning, and sim-to-real robustness training. These developments establish motion execution and balance as important foundations for physical interaction.

Whole-body reference tracking enables rich motion execution~\cite{beyondmimic,gmt} and teleoperated imitation~\cite{twist}. Related work extends expressive control, simulation-to-reality alignment, and motion coverage~\cite{exbody2,asap,unitracker}. OmniH2O uses kinematic poses as a common interface for teleoperation and learned autonomy~\cite{omnih2o}, while TWIST2 supports portable whole-body teleoperation and demonstration collection~\cite{twist2}. AMO combines RL with trajectory optimization to improve stability and expand the executable motion range~\cite{amo}. For loco-manipulation, HOMIE~\cite{homie} combines lower-body control with exoskeleton-mediated upper-body commands. HOVER~\cite{hover} supports multiple control interfaces through policy distillation. Together, these systems provide a range of ways to specify desired humanoid behavior. CompliantBot studies how compliance training can be integrated with such task formulations while preserving the original motion reference or velocity command, so that external interaction can change the executed motion without replacing the task input.

\subsection{Compliant interaction and control}
Classical interaction control specifies a relationship between force and motion. Impedance control regulates mechanical response around a desired motion~\cite{hogan1985impedance}, whereas admittance control converts an external force into a motion response; Ott et al.~\cite{ott2015hybrid} analyze their complementary properties within a common control framework.

Model predictive control provides another component for coordinating whole-body motion. It repeatedly predicts the consequences of candidate actuator commands over a finite horizon, applies the first control action, and replans from the updated state. Howell et al.~\cite{howell2022mjpc} introduce predictive sampling in MuJoCo as a general framework for real-time behavior synthesis. An admittance layer can act upstream of such a controller: the measured force changes a motion reference, and the predictive controller computes the actuation needed to follow it. The nominal task objective and the force-to-motion mapping therefore remain distinct design components. This arrangement contrasts with learning the interaction response directly through training rewards. In the latter case, the policy must learn how to coordinate body motion with task execution, rather than receiving the desired displacement from a separate admittance model.

Learning-based manipulation also builds on explicit force--motion control. Mart\'in-Mart\'in et al.~\cite{martin2019vices} study an end-effector action space with variable impedance, and Beltran-Hernandez et al.~\cite{beltran2020force} combine RL with parallel position/force control or admittance control for rigid position-controlled manipulators. These approaches embed the desired interaction dynamics in the control structure or action representation. For a humanoid, the resulting motion must additionally be coordinated with balance and ground contact, motivating the study of interaction responses involving both local articulation and whole-body movement.

A related line of work trains legged robots to execute tasks under substantial external loads. FALCON~\cite{falcon} uses a force curriculum to maintain manipulation accuracy and locomotion stability. Thor~\cite{thor} coordinates lower-body, waist, and upper-body policies and uses a force-adaptive torso objective to improve whole-body behavior during forceful loco-manipulation. Their emphasis on completing tasks while sustaining interaction forces differs from yielding to reduce a guiding pull. Learned compliant control also has precedents beyond disturbance rejection: Lee et al.~\cite{lee2022deep} use a two-level architecture for simulated humanoids that combines motion imitation with reduced interaction forces and control torques. For legged robots, Hartmann et al.~\cite{deepcompliancego1} relax velocity-tracking incentives during a dedicated push-recovery stage and restore them afterward. These studies show that the interaction response depends on the learning objective as well as the force exposure. The issue is how to coordinate yielding with the task that the controller is already trained to perform.

Other approaches express interaction goals through explicit commands or compliant target motions. Portela et al.~\cite{portela2024} use spring--damper interactions to train end-effector force control policies, adopting separate position and force modes and disabling end-effector position tracking in force mode. Unified force--position control~\cite{unifp} learns responses to both command types for legged loco-manipulation. In reference-based control, SoftMimic~\cite{softmimic} augments whole-body motions using inverse kinematics: the policy observes the original reference but is rewarded for matching a compliant target. GentleHumanoid~\cite{gentlehumanoid} and FACET~\cite{FACET} generate compliant references from impedance dynamics, using multi-link impedance and virtual mass--spring--damper models, respectively. In these formulations, explicit interaction commands or modified reference trajectories specify the desired response before the policy executes it.

CompliantBot takes the original task rewards as the integration point for compliance training. Its state-dependent interaction force field makes the realized load depend on the robot's motion, while a force-magnitude objective evaluates the resulting interaction. Interaction-stage weighting permits temporary task deviations during force application and restores task emphasis after removal. This combination retains the controller's motion reference or velocity command and learns the coordinated response without constructing an additional compliant target trajectory. It provides a common training mechanism for the two task interfaces considered here, with separately trained policies retaining their original inputs and action representations.

\section{Method}
\subsection{Problem formulation}
The humanoid control task is specified by a command $\vect{c}_t$, which is either a whole-body motion reference $\vect{m}_t$ or a velocity command $(v_x^{\rm cmd},v_y^{\rm cmd},\omega_z^{\rm cmd})$. The objective is to preserve nominal task execution, allow temporary deviations during physical interaction, and resume task tracking after the external force is removed.

Let $s_t$ denote the simulator state and $o_t$ the actor observation containing task inputs and robot observations. For each task interface, we train a separate policy conditioned on a finite history $h_t=(o_{t-H+1},\ldots,o_t)$ of length $H$. The policy samples joint-position actions $\vect{a}_t\sim\pi_\theta(\cdot\mid h_t)$, which are executed by the host's native PD actuators with fixed gains. The simulator evolves under these actions and the external interaction force, returning a scalar reward $R_t$ at each control step. The policy parameters $\theta$ are optimized using PPO~\cite{schulman2017ppo} to maximize the expected discounted return,
\begin{equation}
 J(\theta)=\mathbb{E}_{\pi_\theta,\mathcal{D}}
       \left[\sum_{t=0}^{T-1}\gamma^t R_t\right],
 \qquad \gamma=0.99,
\end{equation}
where $T$ is the episode horizon, $\gamma$ is the discount factor, and $\mathcal{D}$ denotes the distribution over task and state initialization, native domain randomization, and the sampled interaction process.

\begin{figure*}[t]
\centering\includegraphics[width=\textwidth,trim=6bp 6bp 6bp 6bp,clip]{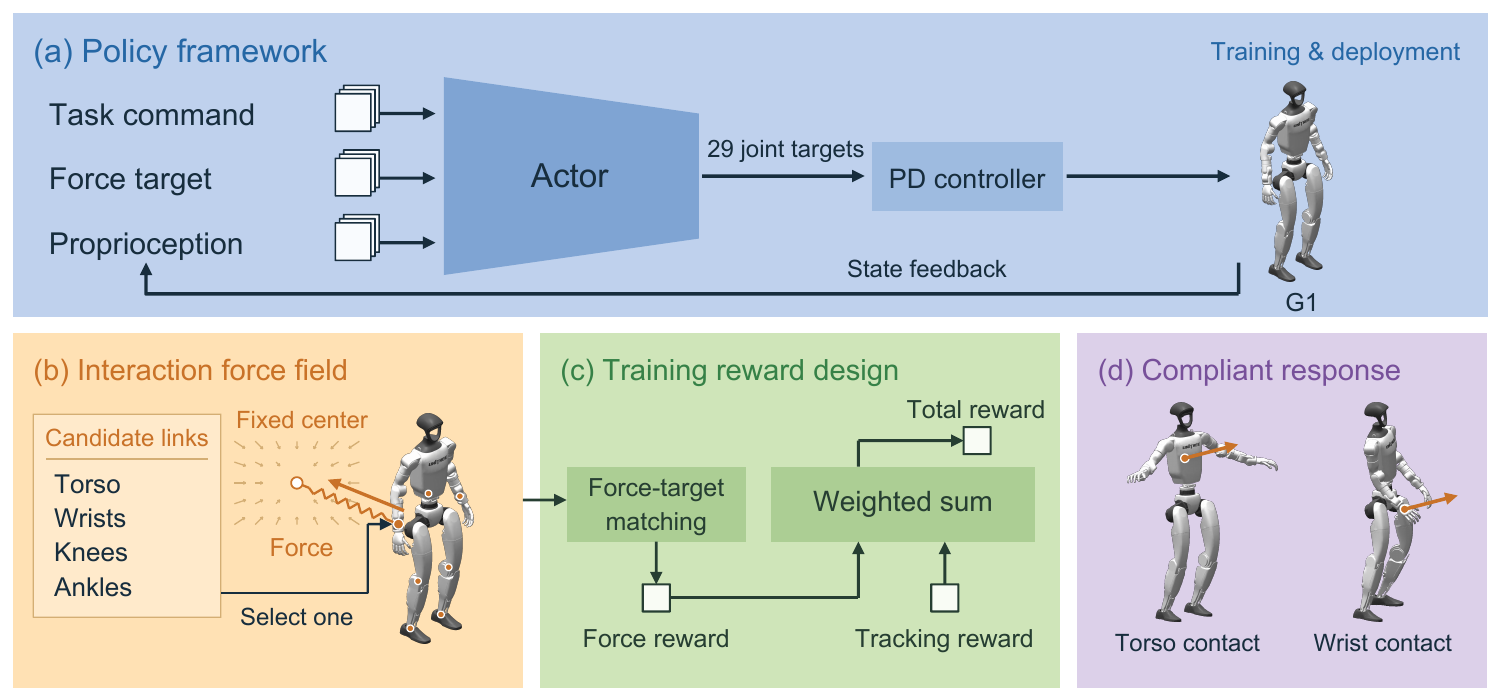}
\caption{\textbf{CompliantBot framework.} (a) The actor generates joint-position targets for a PD controller. (b) A state-dependent interaction force field acts on a sampled body link. (c) The force objective and task rewards are weighted according to the interaction stage. (d) Torso and wrist responses; orange arrows indicate external forces.}
\label{fig:overview}
\end{figure*}

\subsection{State-dependent interaction force field}
To train coordinated responses to forces at different body sites, we construct a state-dependent interaction force field acting on a sampled body link. The field couples the interaction load to the link position, allowing the robot's motion to change the realized force. This motion--load feedback supports learning responses through local articulation, whole-body displacement, or both, without prescribing a compliant target pose or trajectory.

We implement the interaction force field as a virtual spring. Let $\mathcal{B}_F$ be a set of candidate interaction links spanning the torso and upper and lower limbs. At each interaction onset, we sample one link $b\in\mathcal{B}_F$ and place a spring center $\vect{x}_{\mathrm{c}}$ near its current position. The link selection is unchanged and the center is fixed in world coordinates throughout the interaction. Let $\vect{x}_{b,t}$ denote the link's world position and $\chi_t\in\{0,1\}$ indicate whether the external force is active (the loading stage). The applied force is
\begin{equation}
 \vect{F}_t=\chi_t\,
 \frac{kF_{\max}(\vect{x}_{\mathrm{c}}-\vect{x}_{b,t})}
 {\max\!\left(F_{\max},\,k\|\vect{x}_{\mathrm{c}}-\vect{x}_{b,t}\|_2\right)},
 \label{eq:spring}
\end{equation}
where $k>0$ is the external spring stiffness and $F_{\max}>0$ is the applied-force magnitude cap.

During loading, the force points toward the spring center. When the force magnitude is below $F_{\max}$, moving toward the center reduces the force, while moving away increases it. The cap limits the training load at larger position offsets. This state-dependent relationship allows the force-magnitude objective in the next subsection to evaluate the interaction response without specifying which joint or whole-body motions should produce yielding.

During training, the interaction site, initial pull direction, and force magnitude are randomized. In episodes with external forces, the robot first executes its task without interference, then experiences a sustained pull, and continues the task after force removal. Some episodes remain unforced throughout to maintain training on the original task.

\subsection{Interaction-stage reward allocation}
The original task objective rewards reducing motion-reference or velocity-command tracking errors, whereas yielding to an external force often requires temporary deviations from these targets. We therefore combine an interaction-response objective with the original task objective and adjust their relative weights according to the interaction stage, coordinating task execution, yielding, and task recovery after the external force is removed.

The force-magnitude objective encourages the robot to regulate the interaction load through its own motion, without prescribing a target pose or trajectory. Let $\vect{f}^{\star}\in\mathbb{R}_{\geq0}^3$ specify the desired magnitudes of the external force along the three coordinate axes. Setting $\vect{f}^{\star}=\vect{0}$ encourages yielding to reduce the interaction load; nonzero entries specify desired force magnitudes along the corresponding axes. The interaction reward increases as the applied force magnitudes approach these targets:
\begin{equation}
 r_c(t)=\chi_t\exp\!\left[-\frac{\sum_{j\in\{x,y,z\}}
        \left|\,|F_{t,j}|-f^{\star}_j\right|}{\beta_c}\right],
 \label{eq:reward}
\end{equation}
where $F_{t,j}$ is the component of the applied external force $\vect{F}_t$ along world axis $j\in\{x,y,z\}$ at time $t$. The parameter $\beta_c>0$ sets sensitivity to force-magnitude discrepancies, and $\chi_t$ indicates the loading stage. The motion--load relationship thus connects the policy's response to the force-magnitude objective.

To coordinate task execution and yielding, we combine the interaction reward with the original task rewards. We denote the task reward group by $r_m$ and collect the remaining reward terms, with their original weights, in $r_{\rm reg}$. The total reward is
\begin{equation}
 r_t=w_m(t)r_m(t)+\lambda_c w_c(t)r_c(t)+r_{\rm reg}(t),
 \label{eq:total}
\end{equation}
where $\lambda_c$ is the base interaction-reward coefficient.

\begin{figure}[t]
\centering\includegraphics[width=0.93\columnwidth]{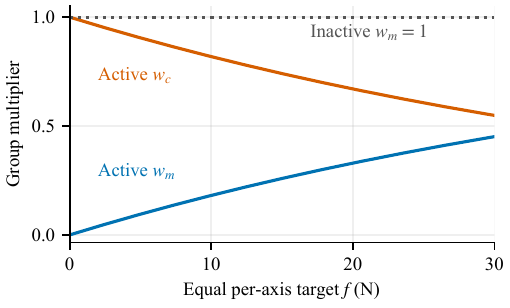}
\caption{\textbf{Interaction-stage reward allocation.} Solid curves show the task and interaction weights under external force; the dotted line shows the task weight during unforced execution.}
\label{fig:weights}
\end{figure}

The interaction and task weights are
\begin{align}
 w_c(t)&=\chi_t\left\langle\exp(-f^{\star}_j/F_s)\right\rangle_{j\in\{x,y,z\}},\\
 w_m(t)&=1-w_c(t).
 \label{eq:weight}
\end{align}
Here $\langle\cdot\rangle_{j\in\{x,y,z\}}$ denotes averaging over the three coordinate axes, and $F_s$ sets the force scale of the weighting rule. Lower force targets assign more weight to interaction and less to the task group, allowing the policy to deviate more from the task while reducing the load; higher targets retain more task weight (Fig.~\ref{fig:weights}).

During force application, the weighting shifts emphasis from task execution toward the interaction objective. After force removal, the original task weight is restored to encourage the policy to resume its task.

\subsection{Integration across task interfaces}
For each controller, CompliantBot applies the interaction-stage weighting to its task reward group $r_m$ and adds the force-magnitude reward. Motion tracking and velocity-command locomotion use separately trained policies with the same interaction force field and reward-allocation mechanism, while retaining their original task inputs and action representations.

The policy receives its original motion reference or velocity command throughout interaction. Under external forces, the interaction objective encourages yielding while reward terms outside the task group retain their original weights. Once the forces are removed, the original task weight is restored to encourage tracking of the current reference or command. For motion tracking, the reference continues to advance during interaction, so task resumption follows its current phase.

Training uses an actor--critic architecture: the actor receives task inputs and proprioceptive history, while the critic can additionally access privileged simulation state. The applied force and interaction-stage indicator are provided by the simulator for reward computation, not as explicit actor inputs (Fig.~\ref{fig:overview}).

\section{Experiments}
\subsection{Experimental setup}
\label{sec:setup}
All tasks use a 29-DoF Unitree G1 with 50\,Hz joint-position targets executed by PD controllers. Policies are trained in NVIDIA Isaac Sim with 200\,Hz physics, giving four simulation steps per control step.

The actor receives a history of observations $o_t$ containing local proprioceptive inputs and task commands $\vect{c}_t$. For both task interfaces, the local proprioceptive inputs include projected gravity $\vect{g}_t$, base angular velocity $\vect{\omega}_t$, joint-position and velocity offsets from their default values $\vect{q}_t$ and $\dot{\vect{q}}_t$, and the previous action $\vect{a}_{t-1}$.

For motion tracking, $\vect{c}_t=\vect{m}_t$ provides reference joint positions $\vect{q}^{\rm ref}_t$ and velocities $\dot{\vect{q}}^{\rm ref}_t$. The actor also receives the reference pelvis position and orientation expressed in the robot's pelvis frame, together with base linear velocity $\vect{v}_t$ and joint efforts $\vect{\tau}_t$. The orientation feature consists of the first two columns of the relative rotation matrix. The observation history length is $H=10$. For velocity-command locomotion, $\vect{c}_t=(v_x^{\rm cmd},v_y^{\rm cmd},\omega_z^{\rm cmd})$ and $H=5$. This interface requires no reference joint states or reference pelvis pose.

The actor and critic use multilayer perceptrons. Table~\ref{tab:training} summarizes the network architecture and PPO settings for motion tracking. During training, the motion-tracking critic receives privileged body poses and interaction information, including the spring center and force magnitude; the locomotion critic receives base linear velocity as an additional state input. Motion-tracking observation normalizers are learned during training and frozen for evaluation; locomotion retains the official observation scaling without learned normalization. The training process, including periodic validation, takes approximately 5.4--6.0 hours per motion-tracking policy on an RTX 4090.

\begin{table}[t]
\centering\CBtablefont
\begin{threeparttable}
\caption{Network and PPO settings for the motion-tracking comparison.}
\label{tab:training}
\setlength{\tabcolsep}{3pt}
\renewcommand{\arraystretch}{0.97}
\begin{tabular*}{0.9\columnwidth}{@{\extracolsep{\fill}}lr@{}}
\toprule[1pt]
\textbf{Parameter} & \textbf{Value}\\
\midrule[0.5pt]
Actor/critic hidden widths & 512, 256, 128\\
Activation & ELU\\
Observation history length $H$ & 10\\
Parallel environments & 4,096\\
Rollout steps per environment & 24\\
PPO updates & 10,000\\
Random seeds per method & 2\\
Initial learning rate & $10^{-3}$\\
Learning-rate schedule & Adaptive KL\\
Discount factor $\gamma$ & 0.99\\
GAE parameter $\lambda$ & 0.95\\
PPO clipping ratio & 0.2\\
Optimization epochs per update & 5\\
Minibatches per epoch & 4\\
Target KL divergence & 0.01\\
Gradient-norm limit & 1.0\\
Entropy coefficient & $0.005\rightarrow0.001$\\
Entropy-decay update interval & 2,000--6,000\\
\bottomrule[1pt]
\end{tabular*}
\end{threeparttable}
\end{table}

Tables~\ref{tab:reward} and~\ref{tab:locomotion_rewards} list the rewards and base coefficients for motion tracking and velocity-command locomotion, respectively. Both use three reward groups: task rewards encourage tracking of the motion reference or velocity command, the compliance reward encourages the desired interaction load, and regularization rewards support stable and smooth movement. During interaction, task and compliance rewards use the interaction-stage weighting, while regularization weights remain unchanged. After force removal, the original task weights are restored.

\begin{table}[t]
\centering\CBtablefont
\begin{threeparttable}
\caption{Reward functions for motion tracking.}
\label{tab:reward}
\setlength{\tabcolsep}{1.5pt}
\renewcommand{\arraystretch}{1.0}
\begin{tabular*}{\columnwidth}{@{\extracolsep{\fill}}llcr@{}}
\toprule[1pt]
\textbf{Group} & \textbf{Reward} & \textbf{Expression} & \textbf{Weight}\\
\midrule[0.5pt]
Compliance & Force target
 & $\chi_t e^{-\|\,|\vect{F}_t|-\vect{f}^{\star}\|_1/\beta_c}$ & 2.0\\
\addlinespace[1pt]
Task & Anchor pos.
 & $e^{-\|\Delta\vect{p}_a\|_2^2/s_{ap}^2}$ & 0.5\\
 & Anchor ori.
 & $e^{-\theta_a^2/s_{aR}^2}$ & 0.5\\
 & Rel. body pos.
 & $e^{-\langle\|\Delta\vect{p}_b\|_2^2\rangle_b/s_p^2}$ & 1.0\\
 & Rel. body ori.
 & $e^{-\langle\theta_b^2\rangle_b/s_R^2}$ & 1.0\\
 & Body lin. vel.
 & $e^{-\langle\|\Delta\vect{v}_b\|_2^2\rangle_b/s_v^2}$ & 1.0\\
 & Body ang. vel.
 & $e^{-\langle\|\Delta\vect{\omega}_b\|_2^2\rangle_b/s_{\omega}^2}$ & 1.0\\
\addlinespace[1pt]
Regularization & Contacts
 & $\sum_{b\in\mathcal{C}}\mathbb{I}[C_{b,t}>F_{\rm thr}]$ & $-0.1$\\
 & Joint limits
 & $\sum_j\ell_j(q_j)$ & $-10.0$\\
 & Action change
 & $\|\vect{a}_t-\vect{a}_{t-1}\|_2^2$ & $-0.1$\\
\bottomrule[1pt]
\end{tabular*}
\end{threeparttable}
\end{table}

We compare CompliantBot with internal control variants, learned compliance methods, and classical controllers. The internal comparisons use the following variants for both motion tracking and velocity-command locomotion:
\begin{itemize}
\item \textbf{Baseline (BS):} The original task controller, trained without the added interaction force field or compliance reward. Motion tracking uses a BeyondMimic-based controller~\cite{beyondmimic}; locomotion uses the official Unitree RL Lab controller. Each retains its original task rewards and domain randomization.
\item \textbf{Disturbance randomization (DR):} A force-trained baseline using the same interaction force field distribution as CB while retaining the original task rewards.
\item \textbf{CompliantBot (CB):} The corresponding controller trained with the interaction force field, force-magnitude objective, and interaction-stage reward allocation.
\end{itemize}
Within each task, the internal variants share the actor--critic architecture and action representation. The external comparisons use the following learned and classical baselines:
\begin{itemize}
\item \textbf{SoftMimic~\cite{softmimic}:} A released whole-body policy trained with compliant targets constructed through inverse kinematics.
\item \textbf{GentleHumanoid~\cite{gentlehumanoid}:} A released policy that uses multi-link impedance to generate compliant motion references.
\item \textbf{Model predictive control (MPC):} A full-body controller adapted to G1 using the iLQG solver in MuJoCo MPC~\cite{howell2022mjpc}, serving as a classical task-tracking baseline. It controls all 29 actuated joints without a learned policy.
\item \textbf{MPC with admittance (MPC+Adm):} The same MPC controller augmented with a mass--damper--spring admittance model~\cite{ott2015hybrid} that adjusts body and wrist references according to the applied force.
\end{itemize}
SoftMimic and GentleHumanoid are compared with CompliantBot in the motion-imitation experiments, while MPC and MPC+Adm are compared with CompliantBot in the velocity-command locomotion experiments.


\begin{table}[t]
\centering\CBtablefont\fontsize{8}{9.5}\selectfont
\begin{threeparttable}
\caption{Reward functions for velocity-command locomotion.}
\label{tab:locomotion_rewards}
\setlength{\tabcolsep}{1pt}
\renewcommand{\arraystretch}{1.05}
\begin{tabular*}{\columnwidth}{@{\extracolsep{\fill}}llcr@{}}
\toprule[1pt]
\textbf{Group} & \textbf{Reward} & \textbf{Expression} & \textbf{Weight}\\
\midrule[0.5pt]
Compliance & Force target
 & $\chi_t e^{-\|\vect{F}_t\|_1/\beta_c}$ & 2.0\\
\addlinespace[1pt]
Task & Linear vel.
 & $e^{-\|\vect{v}_{xy}^{\rm yaw}-\vect{v}_{xy}^{\rm cmd}\|_2^2/\sigma_v^2}$ & 1.0\\
 & Angular vel.
 & $e^{-(\omega_z^{\rm b}-\omega_z^{\rm cmd})^2/\sigma_\omega^2}$ & 0.5\\
\addlinespace[1pt]
Regularization & Alive
 & $\mathbb{I}[\text{not terminated}]$ & 0.15\\
 & Vertical vel.
 & $(v_z^{\rm b})^2$ & $-2.0$\\
 & Roll/pitch vel.
 & $\|\vect{\omega}_{xy}^{\rm b}\|_2^2$ & $-0.05$\\
 & Joint vel.
 & $\|\dot{\vect{q}}\|_2^2$ & $-0.001$\\
 & Joint accel.
 & $\|\ddot{\vect{q}}\|_2^2$ & $-2.5\!\times\!10^{-7}$\\
 & Action change
 & $\|\vect{a}_t-\vect{a}_{t-1}\|_2^2$ & $-0.05$\\
 & Joint limits
 & $\sum_j\ell_j(q_j)$ & $-5.0$\\
 & Joint power
 & $\sum_j|\tau_j\dot q_j|$ & $-2\!\times\!10^{-5}$\\
 & Arm posture
 & $\sum_{j\in\mathcal{J}_A}|q_j-q_j^0|$ & $-0.1$\\
 & Waist posture
 & $\sum_{j\in\mathcal{J}_W}|q_j-q_j^0|$ & $-1.0$\\
 & Hip posture
 & $\sum_{j\in\mathcal{J}_H}|q_j-q_j^0|$ & $-1.0$\\
 & Upright posture
 & $\|\vect{g}_{xy}^{\rm b}\|_2^2$ & $-5.0$\\
 & Base height
 & $(h-h_0)^2$ & $-10.0$\\
 & Gait contact
 & $\eta_t\sum_f\mathbb{I}[c_f=s_f]$ & 0.5\\
 & Foot sliding
 & $\sum_f\mathbb{I}[C_f>1\,\mathrm{N}]\|\vect{v}_{f,xy}^{\rm w}\|_2$ & $-0.2$\\
 & Foot clearance
 & $e^{-\sum_f e_{f,h}^2u_f/\beta_h}$ & 1.0\\
 & Contacts
 & $\sum_{b\in\mathcal{C}}\mathbb{I}[C_{b,t}>1\,\mathrm{N}]$ & $-1.0$\\
\bottomrule[1pt]
\end{tabular*}

\end{threeparttable}
\end{table}

\subsection{Whole-body motion-tracking performance}
\label{sec:performance}
This section evaluates CompliantBot on whole-body motion-tracking tasks, focusing on yielding to external forces, task resumption after force removal, nominal tracking accuracy, and stability. We compare CB with BS, DR, and released SoftMimic and GentleHumanoid policies using the same references.

The evaluation panel contains 224 scenarios per policy: two start-frame repetitions, seven sites, eight horizontal directions, and two force magnitudes (5 and 10\,N). Each rollout lasts 10\,s, with the external force applied continuously from 2 to 5\,s. Yielding is averaged over the final second of force application (4--5\,s), while tracking error after force removal is evaluated over 5--10\,s. We evaluate a no-force control, prescribed constant forces, and capped virtual springs with stiffnesses of $k=20$ and 50\,N/m. CB uses a zero force-magnitude target. Policy parameters and observation normalizers remain frozen during evaluation.

We evaluate stable execution, yielding, tracking after force removal, and the actual interaction load. Survival measures whether a policy completes the trial. World-frame yielding ($Y_w$) measures the extent of the response to external force and is defined as the mean displacement of the pulled body link along the applied force relative to the same policy's unforced rollout. $\Delta E$ denotes the additional tracking error following an external-force perturbation. It is the mean body-position tracking error over 5--10\,s in the forced rollout minus that in the same policy's unforced counterpart. For spring trials, $\bar F$ is the mean actual force magnitude during force application (2--5\,s), used to compare the interaction loads experienced by different policies. Survival is calculated over all test scenarios; the remaining metrics use paired forced and unforced rollouts that both complete the trial.

\begin{table*}[t]
\centering\CBtablefont\fontsize{8}{9.5}\selectfont
\begin{threeparttable}
\caption{Motion-tracking performance under external forces. Bold and underlined means indicate the best and second-best values within each condition, respectively.}
\label{tab:main}
\setlength{\tabcolsep}{0.6pt}
\renewcommand{\arraystretch}{1.4}
\begin{tabular*}{\textwidth}{@{\extracolsep{\fill}}l*{3}{c}@{\hspace{4pt}}*{4}{c}@{\hspace{4pt}}*{4}{c}@{}}
\toprule[1pt]
& \multicolumn{3}{c}{Constant force} & \multicolumn{4}{c}{Spring ($k=20$\,N/m)} & \multicolumn{4}{c}{Spring ($k=50$\,N/m)}\\
\cmidrule(lr){2-4}\cmidrule(lr){5-8}\cmidrule(lr){9-12}
Method & Survival (\%)$\uparrow$ & $Y_w$ (cm)$\uparrow$ & $\Delta E$ (mm)$\downarrow$ & Survival (\%)$\uparrow$ & $Y_w$ (cm)$\uparrow$ & $\Delta E$ (mm)$\downarrow$ & $\bar F$ (N)$\downarrow$ & Survival (\%)$\uparrow$ & $Y_w$ (cm)$\uparrow$ & $\Delta E$ (mm)$\downarrow$ & $\bar F$ (N)$\downarrow$\\
\midrule[0.5pt]
BS & \underline{97.54}$\pm$0.32 & 9.96$\pm$0.30 & 0.58$\pm$0.10 & \underline{97.54}$\pm$0.32 & 2.54$\pm$0.07 & 0.63$\pm$0.03 & 7.04$\pm$0.02 & \underline{97.54}$\pm$0.32 & 0.86$\pm$0.04 & 0.64$\pm$0.06 & 7.27$\pm$0.02\\
DR & 97.32$\pm$0.63 & 6.80$\pm$0.12 & 0.24$\pm$0.03 & 97.32$\pm$0.63 & 1.57$\pm$0.04 & 0.35$\pm$0.02 & 7.06$\pm$0.01 & 97.32$\pm$0.63 & 0.39$\pm$0.13 & 0.34$\pm$0.03 & 7.29$\pm$0.02\\
CB & \textbf{99.78}$\pm$0.32 & \underline{18.75}$\pm$0.29 & \textbf{0.19}$\pm$0.03 & \textbf{99.78}$\pm$0.32 & \underline{10.25}$\pm$0.35 & \textbf{$-$0.10}$\pm$0.50 & \textbf{6.56}$\pm$0.07 & \textbf{99.78}$\pm$0.32 & \textbf{9.48}$\pm$0.46 & \textbf{0.00}$\pm$0.49 & \textbf{6.75}$\pm$0.03\\
SoftMimic & 90.89$\pm$0.40 & \textbf{21.69}$\pm$1.42 & 0.38$\pm$0.44 & 91.43$\pm$2.01 & \textbf{11.48}$\pm$0.87 & 0.40$\pm$0.42 & 6.95$\pm$0.06 & 90.62$\pm$2.07 & \underline{6.73}$\pm$2.05 & 0.41$\pm$0.33 & 7.27$\pm$0.04\\
GentleHumanoid & 95.27$\pm$1.16 & 13.42$\pm$0.52 & \underline{0.22}$\pm$0.04 & 95.18$\pm$0.97 & 6.52$\pm$0.24 & \underline{0.14}$\pm$0.02 & \underline{6.89}$\pm$0.03 & 95.36$\pm$1.08 & 3.74$\pm$0.31 & \underline{0.16}$\pm$0.03 & \underline{7.24}$\pm$0.02\\
\bottomrule[1pt]
\end{tabular*}
\end{threeparttable}
\end{table*}

Table~\ref{tab:main} shows that CB yields more than BS and DR under all three force conditions, whereas DR yields less than BS. This indicates that exposure to external forces alone is insufficient to produce compliant responses. CB also maintains higher survival and smaller additional tracking error after force removal, while reducing the actual interaction load under both spring stiffnesses. Together, these results demonstrate the effectiveness of combining the interaction force field with the compliance reward and interaction-stage reward allocation: the combined mechanism promotes yielding while preserving stable task execution and task recovery.

Compared with the external methods, CB achieves higher survival and smaller additional tracking error after force removal under all three conditions, together with the lowest actual interaction load under both spring stiffnesses. Although SoftMimic yields farther under constant forces and the lower-stiffness spring, its greater displacement under the latter does not translate into a lower interaction load than CB's. Thus, greater yielding displacement does not necessarily imply greater load reduction. CB also yields farther than SoftMimic under the higher-stiffness spring and farther than GentleHumanoid under all three conditions. These results show that CB combines substantial yielding with more effective spring-load reduction, achieving a better overall balance among compliant response, stability, and task recovery.

\begin{figure}[t]
\centering\includegraphics[width=\columnwidth]{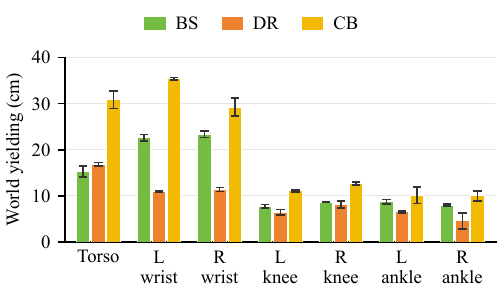}
\caption{\textbf{Yielding across body sites.} Bars and error bars show means and sample standard deviations across training seeds, respectively. Yielding is the displacement of the loaded link relative to the same policy's unforced rollout, projected onto the force direction.}
\label{fig:sites}
\end{figure}

Figure~\ref{fig:sites} compares the yielding responses to 10\,N constant forces applied at different body sites. CB achieves greater mean yielding than BS and DR at every tested site, with larger gains at the torso and wrists; improvements are also observed at the knees and ankles. These results show that the compliance-training approach promotes yielding not only at the upper limbs but also at the torso and lower limbs.

\begin{figure*}[t]
\centering
\includegraphics[width=0.93\textwidth]{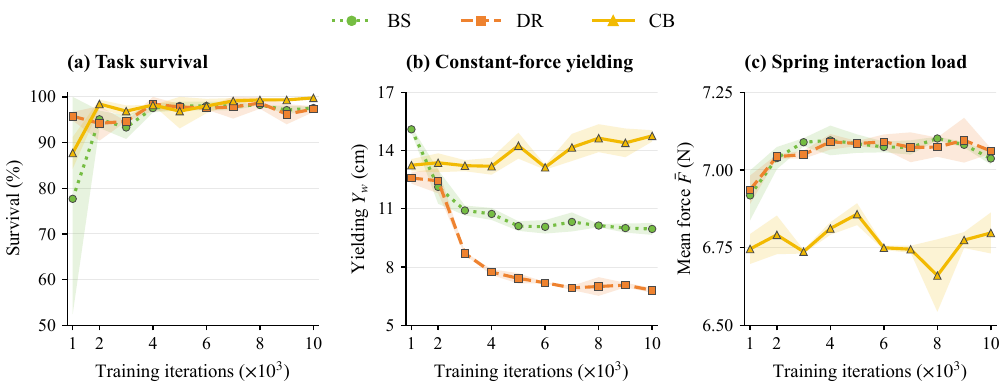}
\caption{\textbf{Learning curves of task survival and compliance.} (a) Task survival and (b) yielding under constant forces; (c) mean interaction force under a spring. Lines and shaded bands show means and sample standard deviations across training seeds, respectively. CB achieves high task survival while retaining substantial yielding and lower spring interaction loads.}
\label{fig:learning_dynamics}
\end{figure*}

The learning curves in Figure~\ref{fig:learning_dynamics} examine how stable execution and compliant responses develop during training. We use the saved validation results at every 1,000 iterations from 1,000 to 10,000 for BS, DR, and CB. Every checkpoint is evaluated on the same scenario grid used in Table~\ref{tab:main}, with the same reference starts, body sites, pull directions, and force levels. The three panels show task survival and $Y_w$ under prescribed constant forces, followed by $\bar F$ under the $k=20$\,N/m spring. Lines and bands indicate the mean and sample standard deviation across training seeds.

Task survival increases overall for BS and CB. DR already achieves high survival at the first evaluation. In the later checkpoints, all methods complete most scenarios, with CB maintaining the highest mean survival over the final four evaluations. These results indicate that compliance training does not compromise the policy's ability to learn reliable task execution.

The yielding curves reveal a different evolution of interaction behavior. BS and DR show progressively smaller force-aligned displacements overall, with DR becoming the least yielding method in the later part of training. CB instead retains a larger response and shows an overall increase toward the end of training, although individual checkpoints fluctuate. From the second evaluated checkpoint onward, its mean yielding exceeds both controls. The joint evolution of survival and yielding indicates that reliable task execution does not require all three methods to learn the same response to an external pull. Retaining the original task objective produces increasingly small deviations, whereas the compliance objective and interaction-stage reward allocation allow substantial yielding to remain part of the learned behavior.

The spring-load curves provide a complementary view. CB has a lower mean actual interaction force than BS and DR at every evaluated checkpoint. This separation is present early and persists through the later checkpoints, while the two controls settle at similar force levels. CB's load fluctuates rather than decreasing monotonically, even as its constant-force yielding increases overall. The two measurements capture different responses: one records displacement under a prescribed force, and the other records the load generated by a state-dependent spring. These results show that CB learns to reduce interaction loads and maintains this ability as training progresses.

\begin{figure}[t]
\centering
\includegraphics[width=\columnwidth]{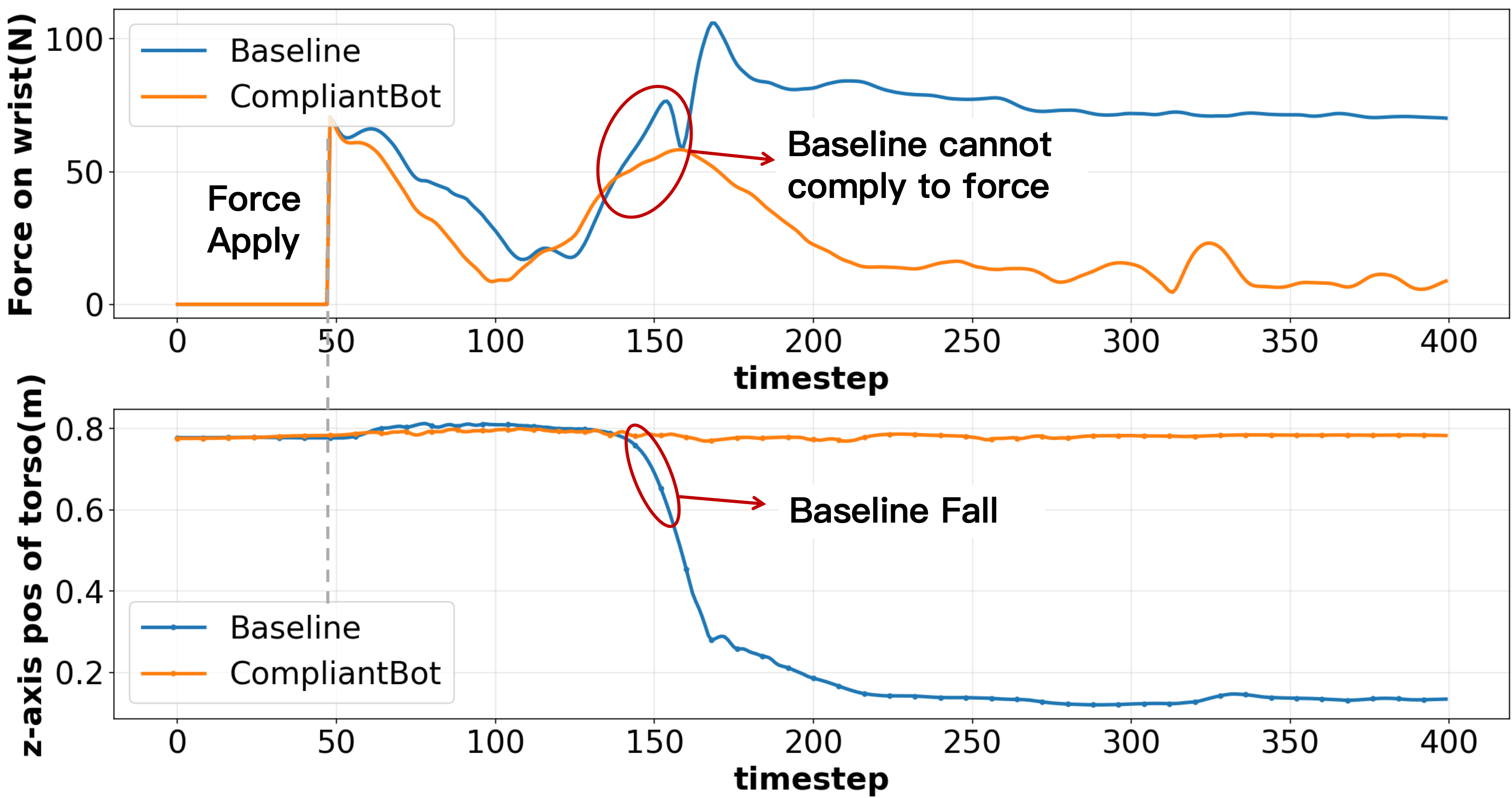}
\caption{\textbf{Interaction force and torso height during dance imitation.} Under a sustained wrist spring pull, the baseline's resistance increases the interaction force and leads to a fall, while CompliantBot yields to reduce the load and maintain stability.}
\label{fig:dance_force_case}
\end{figure}

To illustrate how compliant responses help the robot remain stable and continue its task under external forces, we compare the interaction force and torso height of CB and BS during whole-body dance tracking. Figure~\ref{fig:dance_force_case} shows their responses to a sustained spring pull at the wrist.

As shown in Figure~\ref{fig:dance_force_case}, BS continues to resist the spring pull, causing the interaction force to increase. When the force becomes too large, the robot loses stability and falls, as indicated by the drop in torso height. CB yields to the pull, reducing the interaction force while maintaining its torso height. This case illustrates how compliance can help the robot remain stable and continue a dynamic task by reducing the interaction load.

\subsection{Extension to velocity-command locomotion}
\label{sec:locomotion}
To test whether the compliance-training mechanism extends from motion tracking to velocity-command locomotion, we evaluate yielding, interaction loads, and speed-tracking recovery after force removal while keeping the original velocity-command interface unchanged. We compare BS, DR, and CB using the Unitree RL Lab locomotion controller, with MPC and MPC+Adm as classical references, to examine how these methods balance compliant responses with walking-task execution.

Each rollout uses a body-frame forward-velocity command of 0.5\,m/s and lasts 20\,s, with force applied to the right wrist during 2--5\,s. Loading conditions comprise constant forces of 5 and 10\,N along world $+y$ and a spring with a world-fixed anchor, an initially lateral pull, and stiffness $k=20$\,N/m. Each forced rollout is paired with an unforced rollout of the same policy. We retain the yielding, mean actual force, and survival metrics from Section~\ref{sec:performance}, with survival requiring completion of the 20\,s episode. Yielding $Y_w$ is the mean wrist displacement over 4--5\,s relative to the unforced rollout, projected on the initial force direction; $\bar F$ averages the actual force magnitude over 2--5\,s. Recovery is measured by $\Delta E_v$, the mean absolute forward-speed tracking error over 5--10\,s in the forced rollout minus that in its unforced counterpart.

BS, DR, and CB are evaluated with three evaluation seeds and 64 paired trials per loading condition and seed, retaining native randomization and matched initial states across methods. Table~\ref{tab:locomotion} reports the mean and sample standard deviation across the three per-seed results for each fixed policy.

MPC and MPC+Adm are evaluated on the same scale as BS, DR, and CB, using the same velocity command, loading conditions, and measurement windows. Their closed-loop evaluations use MuJoCo. In MPC+Adm, the external force drives a Cartesian reference offset $\delta\vect{x}$ through the mass--damper--spring model~\cite{ott2015hybrid}
\begin{equation}
M_a\delta\ddot{\vect{x}}+D_a\delta\dot{\vect{x}}+K_a\delta\vect{x}=\vect{F}_t.
\label{eq:baseline_admittance}
\end{equation}
Here $M_a=4$\,kg, $D_a=30$\,N\,s/m, and $K_a=50$\,N/m set the reference dynamics. For the wrist interaction, the reference adjustment combines horizontal whole-body translation with arm inverse kinematics. The adjusted positions and velocities are passed to the same full-body MPC, which determines the actuator commands. Once the applied force is removed, the damped offset returns toward zero, restoring the nominal reference. The force input is the current applied simulation force; neither classical baseline uses a learned policy. This construction separates the predictive task controller from the explicit force-to-motion mapping added by admittance.

\begin{table*}[t]
\centering\CBtablefont\fontsize{8}{9.5}\selectfont
\begin{threeparttable}
\caption{Velocity-command locomotion under external forces. Bold and underlined means indicate the best and second-best values among all five methods within each condition, respectively. Fully tied columns are unmarked.}
\label{tab:locomotion}
\setlength{\tabcolsep}{1pt}
\renewcommand{\arraystretch}{1.05}
\let\CBoriginalshortstack\shortstack
\renewcommand{\shortstack}[1]{\begingroup\setlength{\baselineskip}{8pt}\lineskip=0.3pt\CBoriginalshortstack{#1}\endgroup}
\begin{tabular*}{\textwidth}{@{\extracolsep{\fill}}l*{4}{c}@{\hspace{5pt}}*{4}{c}@{\hspace{5pt}}*{4}{c}@{}}
\toprule[1pt]
& \multicolumn{4}{c}{Constant force (5\,N)} & \multicolumn{4}{c}{Constant force (10\,N)} & \multicolumn{4}{c}{Spring ($k=20$\,N/m)}\\
\cmidrule(lr){2-5}\cmidrule(lr){6-9}\cmidrule(lr){10-13}
Method & \shortstack{$Y_w$\\(cm)$\uparrow$} & \shortstack{$\Delta E_v$\\(mm/s)$\downarrow$} & \shortstack{$\bar F$\\(N)$\downarrow$} & \shortstack{Survival\\(\%)$\uparrow$} & \shortstack{$Y_w$\\(cm)$\uparrow$} & \shortstack{$\Delta E_v$\\(mm/s)$\downarrow$} & \shortstack{$\bar F$\\(N)$\downarrow$} & \shortstack{Survival\\(\%)$\uparrow$} & \shortstack{$Y_w$\\(cm)$\uparrow$} & \shortstack{$\Delta E_v$\\(mm/s)$\downarrow$} & \shortstack{$\bar F$\\(N)$\downarrow$} & \shortstack{Survival\\(\%)$\uparrow$}\\
\midrule[0.5pt]
\raisebox{4pt}{BS} & \shortstack{$\underline{13.75}$\\$\pm 0.04$} & \shortstack{$2.85$\\$\pm 0.39$} & \shortstack{$5.000$\\$\pm 0.000$} & \raisebox{4pt}{$\mathbf{100}$} & \shortstack{$\underline{27.10}$\\$\pm 0.06$} & \shortstack{$1.82$\\$\pm 0.28$} & \shortstack{$10.000$\\$\pm 0.000$} & \raisebox{4pt}{$\mathbf{100}$} & \shortstack{$14.82$\\$\pm 0.12$} & \shortstack{$21.33$\\$\pm 0.29$} & \shortstack{$\underline{13.485}$\\$\pm 0.006$} & \raisebox{4pt}{$\mathbf{100}$}\\
\raisebox{4pt}{DR} & \shortstack{$9.85$\\$\pm 0.08$} & \shortstack{$-0.02$\\$\pm 0.10$} & \shortstack{$5.000$\\$\pm 0.000$} & \raisebox{4pt}{$\mathbf{100}$} & \shortstack{$20.03$\\$\pm 0.06$} & \shortstack{$\mathbf{-0.27}$\\$\pm 0.09$} & \shortstack{$10.000$\\$\pm 0.000$} & \raisebox{4pt}{$\mathbf{100}$} & \shortstack{$\underline{16.90}$\\$\pm 0.05$} & \shortstack{$\underline{10.30}$\\$\pm 0.15$} & \shortstack{$14.366$\\$\pm 0.001$} & \raisebox{4pt}{$\mathbf{100}$}\\
\raisebox{4pt}{CB} & \shortstack{$\mathbf{17.62}$\\$\pm 0.08$} & \shortstack{$\underline{-0.06}$\\$\pm 0.04$} & \shortstack{$5.000$\\$\pm 0.000$} & \raisebox{4pt}{$\mathbf{100}$} & \shortstack{$\mathbf{39.38}$\\$\pm 0.50$} & \shortstack{$\underline{1.64}$\\$\pm 0.49$} & \shortstack{$10.000$\\$\pm 0.000$} & \raisebox{4pt}{$\mathbf{100}$} & \shortstack{$\mathbf{19.78}$\\$\pm 0.18$} & \shortstack{$\mathbf{9.55}$\\$\pm 1.10$} & \shortstack{$\mathbf{10.283}$\\$\pm 0.010$} & \raisebox{4pt}{$\mathbf{100}$}\\
\raisebox{4pt}{MPC} & \shortstack{$1.49$\\$\pm 2.26$} & \shortstack{$1.40$\\$\pm 4.32$} & \shortstack{$5.000$\\$\pm 0.000$} & \raisebox{4pt}{$\underline{75}$} & \shortstack{$2.26$\\$\pm 2.25$} & \shortstack{$12.77$\\$\pm 8.37$} & \shortstack{$10.000$\\$\pm 0.000$} & \raisebox{4pt}{$\underline{75}$} & \shortstack{$2.22$\\$\pm 2.35$} & \shortstack{$15.15$\\$\pm 8.87$} & \shortstack{$15.714$\\$\pm 0.082$} & \raisebox{4pt}{$\underline{75}$}\\
\raisebox{4pt}{MPC+Adm} & \shortstack{$11.52$\\$\pm 2.18$} & \shortstack{$\mathbf{-0.82}$\\$\pm 4.44$} & \shortstack{$5.000$\\$\pm 0.000$} & \raisebox{4pt}{$\underline{75}$} & \shortstack{$18.92$\\$\pm 1.79$} & \shortstack{$31.80$\\$\pm 67.75$} & \shortstack{$10.000$\\$\pm 0.000$} & \raisebox{4pt}{$50$} & \shortstack{$15.62$\\$\pm 2.00$} & \shortstack{$39.80$\\$\pm 48.68$} & \shortstack{$13.964$\\$\pm 0.074$} & \raisebox{4pt}{$50$}\\
\bottomrule[1pt]
\end{tabular*}
\end{threeparttable}
\end{table*}

Table~\ref{tab:locomotion} shows that CB yields farther than BS and DR under both constant-force conditions, while DR yields less than BS. Under the spring, CB also achieves greater mean yielding and a lower actual interaction load than these controls. All three policies complete every trial. After force removal, CB's additional speed-tracking error is close to zero under constant forces and smaller than BS and DR under the spring. These internal comparisons show that force exposure alone is insufficient to promote yielding. Combining the state-dependent interaction force field, compliance reward, and interaction-stage reward allocation promotes yielding and load reduction during velocity-command locomotion while preserving task completion and speed-tracking recovery after release.

Compared with MPC and MPC+Adm, CB achieves greater mean yielding and higher survival under all tested loading conditions, together with a lower actual load and a smaller post-release speed-tracking error under the spring. Within the classical pair, adding admittance increases yielding and reduces the spring load, but is accompanied by lower survival under the 10\,N pull and the spring, revealing a trade-off between compliant response and stability in this configuration. CB achieves more substantial yielding and more effective unloading while completing every trial, demonstrating a better overall balance among compliant response, stability, and task recovery.

Together with the preceding motion-tracking experiments, these results show that CompliantBot's compliance-training mechanism applies to both motion-reference and velocity-command interfaces. It preserves the original task inputs while enabling the robot to yield to external forces, reduce interaction loads, and resume task tracking after release.

\subsection{Task completion and mechanical power under stronger forces}
\label{sec:completion_power}
To examine task continuation and mechanical effort under stronger external forces, we evaluate BS, DR, CB, SoftMimic, and GentleHumanoid on the same walking reference under prescribed pulls. For the versions evaluated here, BS, DR, CB, and GentleHumanoid use a training force-magnitude cap of 50\,N. The 50\,N test therefore examines performance at this magnitude boundary, whereas the 70\,N test evaluates out-of-distribution generalization in force magnitude. SoftMimic retains the 140\,N peak-force limit used in its compliant motion data generation, so neither test magnitude exceeds that limit.

\begin{figure*}[t]
\centering
\includegraphics[width=\textwidth]{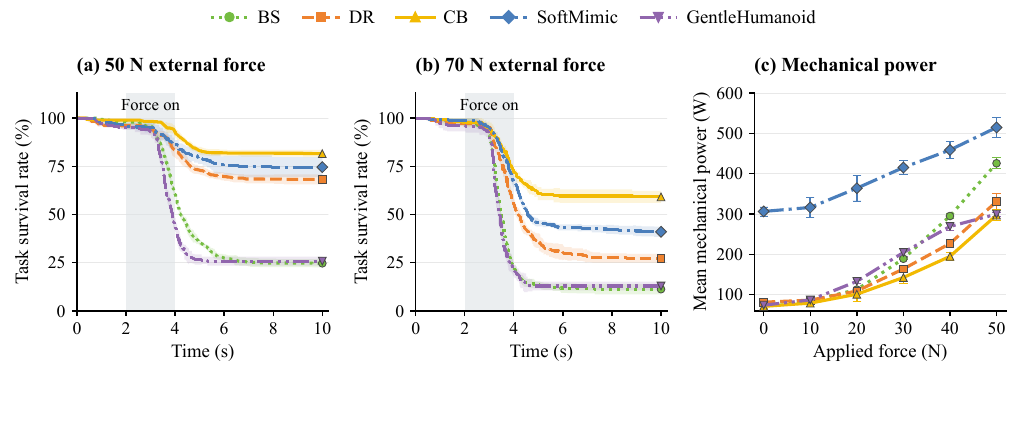}
\caption{\textbf{Task completion and mechanical power under external forces.} (a,b) Task survival rates over time; final values indicate task completion rates. (c) Mean joint mechanical power across force magnitudes. Gray shading marks force application; colored bands and error bars show sample standard deviations across evaluation resets. CB achieves the highest final task completion and the lowest mean mechanical power among the compared methods.}
\label{fig:completion_power}
\end{figure*}

Each trial lasts 10\,s, with force applied from 2 to 4\,s to one body site: the torso, right wrist, or right ankle. Each site is tested in eight horizontal directions with two reference starts per site--direction pair, giving 48 trials per force level and checkpoint. Figure~\ref{fig:completion_power}(a,b) shows the fraction of trials that have not reached a task-termination condition; the final value is the task completion rate. Figure~\ref{fig:completion_power}(c) compares mean mechanical power during unforced execution and under 10--50\,N pulls in increments of 10\,N. Mechanical power is the sum of the absolute joint torque--velocity products, averaged over the force-application interval; unforced trials use the same time window. At each force level and evaluation reset, power is compared across all five methods on common valid case--time samples from their first episodes. All three panels report means and sample standard deviations over five evaluation resets with fixed policies. Within each reset, the two internal checkpoint results are averaged for each method; each external method uses one checkpoint.

CB achieves the highest final completion rate at both force levels, and its advantage persists after force removal, indicating better continuation of the tracking task following interaction. Although completion decreases when the force exceeds the training magnitude cap, CB remains above the other methods, supporting generalization to larger force magnitudes. SoftMimic's training data permit higher peak forces, yet its completion rate remains below CB in this evaluation, indicating that force-magnitude coverage alone does not determine task continuation.

For mechanical power, CB has the lowest mean at every tested force magnitude, including unforced execution, although its margin over each method varies with force magnitude. Together with the completion results, CB maintains both higher task completion and lower mechanical power at 50\,N, while retaining higher task continuation in the 70\,N out-of-distribution test. These experiments show that CB can balance task execution and mechanical effort under stronger interaction loads while extending task retention beyond the training force-magnitude range.

\subsection{Real-world experiments}
\label{sec:real_world}
Real-robot demonstrations examine how the robot coordinates local joint motion and whole-body movement in response to physical interaction. The supplementary video presents the method, simulation comparisons, and recordings of these real-robot demonstrations.

\begin{figure*}[t]
\centering
\includegraphics[width=\textwidth]{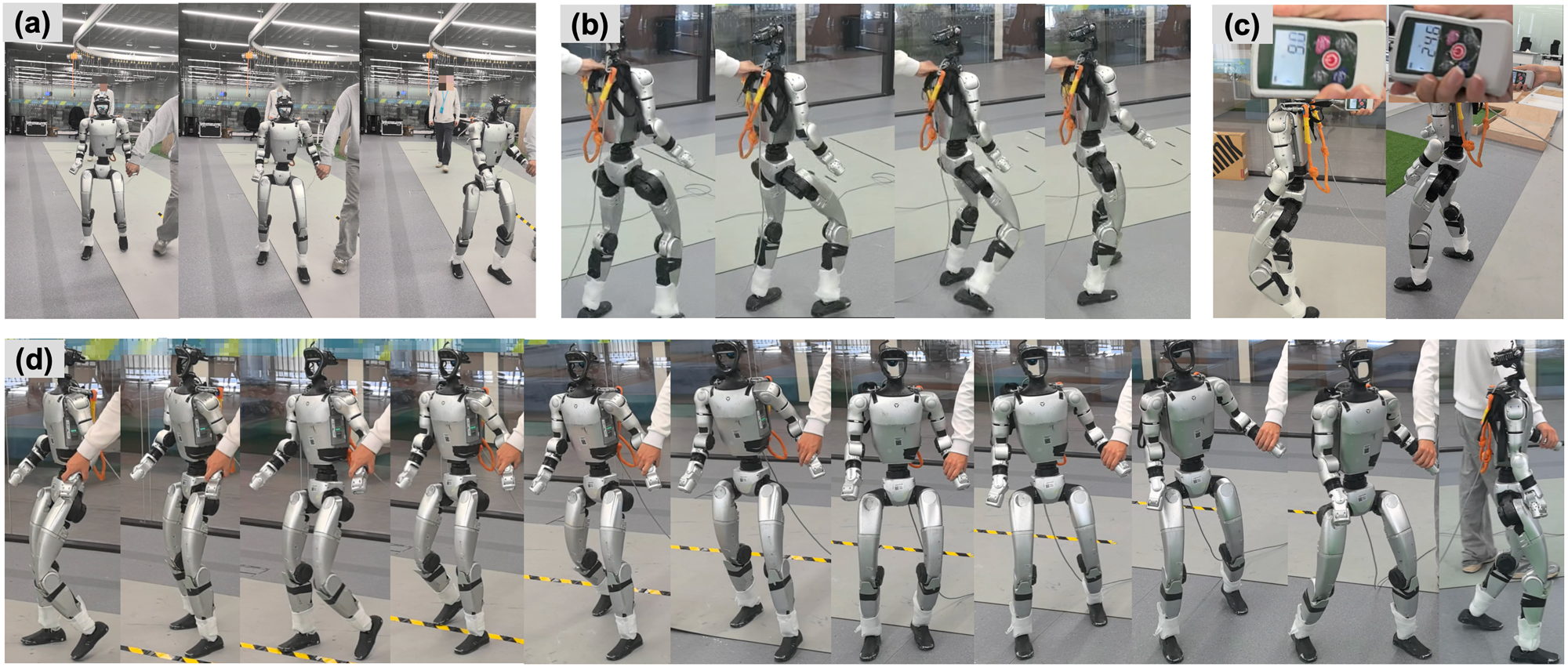}
\caption{\textbf{Real-robot compliant responses during walking.} (a) Arm yielding and lateral stepping under wrist pulls. (b) Stepping in place and backward stepping under torso pulls. (c) Higher force-magnitude targets $\vect{f}^{\star}$ require stronger pulls to stop forward motion. (d) A sustained wrist pull turns the robot around the fixed hand. These demonstrations show that CompliantBot coordinates local and whole-body motion to yield while continuing the task.}
\label{fig:hardware_walk}
\label{fig:hardware}
\end{figure*}

During forward walking, a modest wrist pull causes the robot to raise its arm in the direction of the pull. With a stronger pull, the response also includes lateral stepping, allowing the whole body to follow the interaction rather than relying on arm motion alone (Figure~\ref{fig:hardware_walk}a). A backward torso pull produces stepping in place and backward stepping as the pull increases (Figure~\ref{fig:hardware_walk}b). The robot thus adapts its walking to pulls at different body sites. The real-robot experiments also compare interaction responses under different force-magnitude targets $\vect{f}^{\star}$. A higher target requires a stronger pull to stop the robot from walking forward, leaving it stepping in place (Figure~\ref{fig:hardware_walk}c).

Sustained wrist constraint produces a different response. When the wrist is held near a fixed location, the robot turns and changes its walking direction around the constrained hand (Figure~\ref{fig:hardware_walk}d). Alongside the wrist and torso pulls, this sequence shows how compliant behavior can involve arm articulation, stepping, and body reorientation during an ongoing walking task.

An elastic-band interaction further examines the response during upper-body dance imitation. One end of the band is attached to the wrist while the other is held at a fixed location. Compared with the baseline, CompliantBot moves closer to the fixed end and reduces the amplitude of its arm motion, shortening the band's extension. The real-robot comparison in Figure~\ref{fig:teaser} illustrates the corresponding posture adjustment under the elastic pull.

These real-robot results show that CompliantBot coordinates local and whole-body compliant responses through arm articulation, stepping, and body reorientation, and reduces the elastic band's extension through posture adjustments. This demonstrates its ability to integrate yielding into ongoing walking and motion execution, combining compliant responses with continued task execution during real physical interaction.

\section{Discussion and Conclusion}
The internal comparisons with BS and DR show that force exposure alone is insufficient to promote yielding and load reduction. CompliantBot combines the state-dependent interaction force field, compliance reward, and interaction-stage reward allocation to allow temporary task deviations during force application and reward the robot for motion that brings the actual force magnitude toward its target. The robot can achieve this through local articulation, whole-body displacement, or their coordination, without a prescribed compliant motion trajectory. Greater yielding, lower spring loads, and task recovery after release show that this combined design supports compliant responses alongside continued task execution.

The comparisons with SoftMimic and GentleHumanoid show that greater yielding displacement does not necessarily produce more effective unloading. Although SoftMimic yields farther under the lower-stiffness spring, CB experiences a lower actual load while maintaining higher survival and a smaller post-release tracking error. In velocity-command locomotion, CB achieves greater mean yielding and higher survival than MPC and MPC+Adm across the tested loading conditions, together with a lower interaction load and a smaller post-release speed-tracking error under the spring. These external comparisons show that CompliantBot achieves a better overall balance among compliant response, effective unloading, stability, and task recovery.

The motion-tracking and velocity-command locomotion experiments further demonstrate the applicability of this mechanism across different task interfaces. The former specifies the task through motion references, while the latter uses desired linear and angular velocities. CompliantBot retains the original task inputs in both interfaces, allowing temporary deviations during loading and resuming task tracking after release. These results show that the compliance-training mechanism can be integrated into different task objectives without separately constructing compliant data or target trajectories for each command interface.

The current approach remains limited by its lack of scene perception. Policies respond to external forces through task inputs and proprioceptive history, without explicit information about surrounding obstacles, object geometry, or the purpose of an interaction. The present experiments do not examine how the robot should use scene context to decide when to yield, where to move, and which contacts to resist. Future work could incorporate visual or depth observations to guide compliant motion using environmental constraints and interaction context.

The current evaluations also focus on motion tracking and velocity-command locomotion under external pulls, without complex object manipulation. Carrying, coordinated bimanual handling, and contact-rich assembly require maintaining useful contact forces while satisfying grasp and object-motion constraints; reducing interaction loads alone is insufficient to complete these tasks. Future extensions should combine object-state information with task-dependent contact objectives and evaluate manipulation accuracy, contact-force regulation, and task completion.

CompliantBot combines a state-dependent interaction force field, a force-magnitude objective, and interaction-stage reward allocation to learn yielding and task resumption while preserving the original task inputs, without preconstructed compliant data or target trajectories. Motion-tracking and velocity-command locomotion experiments demonstrate the effectiveness of the combined training mechanism, while external comparisons show a better overall balance among compliant response, stability, and task recovery. Under stronger pulls, CB achieves the highest task completion among the compared methods, retaining this advantage beyond its training force-magnitude range. Power evaluations also show the lowest mean joint mechanical power among the compared methods. Real-robot results further demonstrate coordinated local and whole-body responses that combine compliance with continued task execution during physical interaction.

\bibliographystyle{elsarticle-num}
\balance
\bibliography{references,additional_references,paper34_references}

\begin{thebibliography}{10}
\expandafter\ifx\csname url\endcsname\relax
  \def\url#1{\texttt{#1}}\fi
\expandafter\ifx\csname urlprefix\endcsname\relax\def\urlprefix{URL }\fi
\expandafter\ifx\csname href\endcsname\relax
  \def\href#1#2{#2} \def\path#1{#1}\fi

\bibitem{beyondmimic}
Q.~Liao, T.~E. Truong, X.~Huang, Y.~Gao, G.~Tevet, K.~Sreenath, C.~K. Liu,
  {BeyondMimic}: From motion tracking to versatile humanoid control via guided
  diffusion, arXiv preprint arXiv:2508.08241 (2025).

\bibitem{cassie-loco}
I.~Radosavovic, T.~Xiao, B.~Zhang, T.~Darrell, J.~Malik, K.~Sreenath,
  Real-world humanoid locomotion with reinforcement learning, Science Robotics
  9~(89) (2024) eadi9579.

\bibitem{gmt}
Z.~Chen, M.~Ji, X.~Cheng, X.~Peng, X.~B. Peng, X.~Wang, {GMT}: General motion
  tracking for humanoid whole-body control, arXiv preprint arXiv:2506.14770
  (2025).

\bibitem{twist}
Y.~Ze, Z.~Chen, J.~P. Ara{\'u}jo, Z.-a. Cao, X.~B. Peng, J.~Wu, C.~K. Liu,
  {TWIST}: Teleoperated whole-body imitation system, arXiv preprint
  arXiv:2505.02833 (2025).

\bibitem{homie}
Q.~Ben, F.~Jia, J.~Zeng, J.~Dong, D.~Lin, J.~Pang, {HOMIE}: Humanoid
  loco-manipulation with isomorphic exoskeleton cockpit, arXiv preprint
  arXiv:2502.13013 (2025).

\bibitem{hover}
T.~He, W.~Xiao, T.~Lin, Z.~Luo, Z.~Xu, Z.~Jiang, J.~Kautz, C.~Liu, G.~Shi,
  X.~Wang, L.~Fan, Y.~Zhu, {HOVER}: Versatile neural whole-body controller for
  humanoid robots, in: 2025 IEEE International Conference on Robotics and
  Automation (ICRA), IEEE, 2025, pp. 9989--9996.

\bibitem{softmimic}
G.~B. Margolis, M.~Wang, N.~Fey, P.~Agrawal, {SoftMimic}: Learning compliant
  whole-body control from examples, arXiv preprint arXiv:2510.17792 (2025).

\bibitem{gentlehumanoid}
Q.~Lu, Y.~Feng, B.~Shi, M.~Piseno, Z.~Bao, C.~K. Liu, {GentleHumanoid}:
  Learning upper-body compliance for contact-rich human and object interaction,
  arXiv preprint arXiv:2511.04679 (2025).

\bibitem{deepcompliancego1}
A.~Hartmann, D.~Kang, F.~Zargarbashi, M.~Zamora, S.~Coros, Deep compliant
  control for legged robots, in: 2024 IEEE International Conference on Robotics
  and Automation (ICRA), IEEE, 2024, pp. 11421--11427.

\bibitem{h1-loco}
J.~Long, J.~Ren, M.~Shi, Z.~Wang, T.~Huang, P.~Luo, J.~Pang, Learning humanoid
  locomotion with perceptive internal model, in: 2025 IEEE International
  Conference on Robotics and Automation (ICRA), IEEE, 2025, pp. 9997--10003.

\bibitem{lipschitz}
Z.~Chen, X.~He, Y.-J. Wang, Q.~Liao, Y.~Ze, Z.~Li, S.~S. Sastry, J.~Wu,
  K.~Sreenath, S.~Gupta, X.~B. Peng, Learning smooth humanoid locomotion
  through {Lipschitz}-constrained policies, in: 2025 IEEE/RSJ International
  Conference on Intelligent Robots and Systems (IROS), IEEE, 2025, pp.
  4743--4750.

\bibitem{cassie-ijrr}
Z.~Li, X.~B. Peng, P.~Abbeel, S.~Levine, G.~Berseth, K.~Sreenath, Reinforcement
  learning for versatile, dynamic, and robust bipedal locomotion control, The
  International Journal of Robotics Research 44~(5) (2025) 840--888.
\newblock \href {https://doi.org/10.1177/02783649241285161}
  {\path{doi:10.1177/02783649241285161}}.

\bibitem{hub}
T.~Zhang, B.~Zheng, R.~Nai, Y.~Hu, Y.-J. Wang, G.~Chen, F.~Lin, J.~Li, C.~Hong,
  K.~Sreenath, Y.~Gao, {HuB}: Learning extreme humanoid balance, arXiv preprint
  arXiv:2505.07294 (2025).

\bibitem{exbody2}
M.~Ji, X.~Peng, F.~Liu, J.~Li, G.~Yang, X.~Cheng, X.~Wang, {ExBody2}: Advanced
  expressive humanoid whole-body control, arXiv preprint arXiv:2412.13196
  (2024).

\bibitem{asap}
T.~He, J.~Gao, W.~Xiao, Y.~Zhang, Z.~Wang, J.~Wang, Z.~Luo, G.~He,
  N.~Sobanbabu, C.~Pan, Z.~Yi, G.~Qu, K.~Kitani, J.~Hodgins, L.~Fan, Y.~Zhu,
  C.~Liu, G.~Shi, {ASAP}: Aligning simulation and real-world physics for
  learning agile humanoid whole-body skills, arXiv preprint arXiv:2502.01143
  (2025).

\bibitem{unitracker}
K.~Yin, W.~Zeng, K.~Fan, M.~Dai, Z.~Wang, Q.~Zhang, Z.~Tian, J.~Wang, J.~Pang,
  W.~Zhang, {UniTracker}: Learning universal whole-body motion tracker for
  humanoid robots, arXiv preprint arXiv:2507.07356 (2025).

\bibitem{omnih2o}
T.~He, Z.~Luo, X.~He, W.~Xiao, C.~Zhang, W.~Zhang, K.~Kitani, C.~Liu, G.~Shi,
  {OmniH2O}: Universal and dexterous human-to-humanoid whole-body teleoperation
  and learning, arXiv preprint arXiv:2406.08858 (2024).

\bibitem{twist2}
Y.~Ze, S.~Zhao, W.~Wang, A.~Kanazawa, R.~Duan, P.~Abbeel, G.~Shi, J.~Wu, C.~K.
  Liu, {TWIST2}: Scalable, portable, and holistic humanoid data collection
  system, arXiv preprint arXiv:2511.02832 (2025).

\bibitem{amo}
J.~Li, X.~Cheng, T.~Huang, S.~Yang, R.-Z. Qiu, X.~Wang, {AMO}: Adaptive motion
  optimization for hyper-dexterous humanoid whole-body control, arXiv preprint
  arXiv:2505.03738 (2025).

\bibitem{hogan1985impedance}
N.~Hogan, Impedance control: An approach to manipulation: Part {I}---theory,
  Journal of Dynamic Systems, Measurement, and Control 107~(1) (1985) 1--7.

\bibitem{ott2015hybrid}
C.~Ott, R.~Mukherjee, Y.~Nakamura, A hybrid system framework for unified
  impedance and admittance control, Journal of Intelligent \& Robotic Systems
  78 (2015) 359--375.
\newblock \href {https://doi.org/10.1007/s10846-014-0082-1}
  {\path{doi:10.1007/s10846-014-0082-1}}.

\bibitem{howell2022mjpc}
T.~Howell, N.~Gileadi, S.~Tunyasuvunakool, K.~Zakka, T.~Erez, Y.~Tassa,
  {Predictive Sampling: Real-time Behaviour Synthesis with MuJoCo}, arXiv
  preprint arXiv:2212.00541 (2022).
\newblock \href {https://doi.org/10.48550/arXiv.2212.00541}
  {\path{doi:10.48550/arXiv.2212.00541}}.

\bibitem{martin2019vices}
R.~Mart{\'i}n-Mart{\'i}n, M.~A. Lee, R.~Gardner, S.~Savarese, J.~Bohg, A.~Garg,
  Variable impedance control in end-effector space: An action space for
  reinforcement learning in contact-rich tasks, in: 2019 IEEE/RSJ International
  Conference on Intelligent Robots and Systems (IROS), IEEE, 2019, pp.
  1010--1017.

\bibitem{beltran2020force}
C.~C. Beltran-Hernandez, D.~Petit, I.~G. Ramirez-Alpizar, T.~Nishi, S.~Kikuchi,
  T.~Matsubara, K.~Harada, Learning force control for contact-rich manipulation
  tasks with rigid position-controlled robots, IEEE Robotics and Automation
  Letters 5~(4) (2020) 5709--5716.

\bibitem{falcon}
Y.~Zhang, Y.~Yuan, P.~Gurunath, I.~Gupta, S.~Omidshafiei, A.-a. Agha-mohammadi,
  M.~Vazquez-Chanlatte, L.~Pedersen, T.~He, G.~Shi, {FALCON}: Learning
  force-adaptive humanoid loco-manipulation, arXiv preprint arXiv:2505.06776
  (2025).

\bibitem{thor}
G.~Li, H.~Shi, Q.~Shi, Y.~Hu, C.~Ma, Z.~Wang, X.~Wang, S.~Luo, {Thor}: Towards
  human-inspired whole-body reactions for intense contact-rich environments,
  arXiv preprint arXiv:2510.26280 (2025).

\bibitem{lee2022deep}
S.~Lee, P.~S. Chang, J.~Lee, Deep compliant control, in: ACM SIGGRAPH 2022
  Conference Proceedings, 2022, pp. 1--9.
\newblock \href {https://doi.org/10.1145/3528233.3530719}
  {\path{doi:10.1145/3528233.3530719}}.

\bibitem{portela2024}
T.~Portela, G.~B. Margolis, Y.~Ji, P.~Agrawal, Learning force control for
  legged manipulation, in: IEEE International Conference on Robotics and
  Automation (ICRA), 2024, pp. 15366--15372.

\bibitem{unifp}
P.~Zhi, P.~Li, J.~Yin, B.~Jia, S.~Huang, Learning a unified policy for position
  and force control in legged loco-manipulation, arXiv preprint
  arXiv:2505.20829 (2025).

\bibitem{FACET}
B.~Xu, H.~Weng, Q.~Lu, Y.~Gao, H.~Xu, {FACET}: Force-adaptive control via
  impedance reference tracking for legged robots, arXiv preprint
  arXiv:2505.06883 (2025).

\bibitem{schulman2017ppo}
J.~Schulman, F.~Wolski, P.~Dhariwal, A.~Radford, O.~Klimov, Proximal policy
  optimization algorithms, arXiv preprint arXiv:1707.06347 (2017).

\end{thebibliography}
\end{document}